# CITlab ARGUS for Arabic Handwriting*

Description of CITlab's system for the OpenHaRT 2013 Document Image Recognition task

Gundram Leifert, Roger Labahn, and Tobias Strauß
CITlab, Institute for Mathematics
University of Rostock
Rostock, Germany
{gundram.leifert, roger.labahn, tobias.strauss}@uni-rostock.de

*Abstract*—In recent years, it has been shown that multi-dimensional recurrent neural networks (MDRNN) perform very well in offline handwriting recognition problems like the OpenHaRT 2013 Document Image Recognition (DIR) task. With suitable writing preprocessing and dictionary lookup, our ARGUS software completed this task with an error rate of 26.27% in its primary setup.

## I. Introduction

For the OpenHaRT 2013 evaluation, our core technology is the neural network of [2] which has been further improved by substituting the Multi-Dimensional Long Short-Term Memory (MDLSTM) cells by newly developed Multi-Dimensional Leaky (MDLeaky) cells. Before presenting some writing to the network, we preprocess it by slant and height normalizations. Then directly using the raw output of the network, CITlab's ARGUS software achieves an OpenHaRT 2013 word error rate (WER) of 33.14% on the evaluation dataset. This could be improved down to 24.60% in the best contrastive system setups where we used different dictionaries or lookup strategies for detecting permissible words.

## II. Submissions

The Computational Intelligence Technology Lab (CITlab) team at the University of Rostock submitted various systems that differ slightly in their neural network and, much more important, in the particular dictionary lookup. Our OpenHaRT 2013 SYSID (see [1]) is ARGUS_<network label>_<decoder number>, and it is composed of the software name *ARGUS*, the label *s* or *l* for the two neural network versions, and a certain decoder number which will be explained in detail in III.C and IV.

## III. Primary System Specifications

The engine description is divided into three parts: writing preprocessing, neural network setup and training, and decoding. For all topics related to OpenHART 2013 MADCAT data provided by LDC, we refer to [1] for further explanation and details.

### A. Preprocessing

The given training data consist of scanned pages of handwritten Arabic texts. According to the evaluation data setup, for each line, a certain polygon around the whole line was constructed connecting the given word polygons. This gives the initial line image. While we do not apply any image preprocessing to the scans themselves, we correct the slants and normalize the images to a specific height. In general, that is apart from possible minor modifications at the ends of the writings, the procedure is the following – see Fig. 1 for an example. For a local environment, we calculate the median of the black pixels in vertical direction. The main body of the characters is considered to be 80 pixels above and 60 pixels below this median – see Fig. 1, top, where the dotted curves show the median and the boundaries of the main body. All pixels are now shifted vertically such that the main body becomes horizontal. Then, the remaining parts below and above the main body are shrinked such that a fixed overall height of 180 pixels is achieved – see Fig. 1, middle. This shift generates a certain new slant of the writing against the new baseline, which is compensated for by an appropriate horizontal shift of the pixels. Fig. 1, bottom, shows the completely preprocessed line image which we call *writing* in the following. After a final scaling by factor 0.5, such writings are presented to the network.

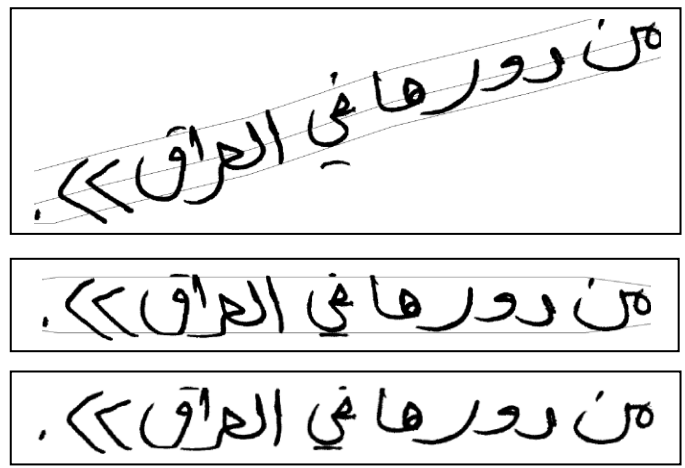

Fig. 1. Preprocessing of a line. Top: Raw image extracted by the given line polygon. Middle: Locally corrected vertical shift and shrink to normalized height. Bottom: Globally corrected horizontal shift.

* Funded by research grant no. V220-630-08-TFMV-S/F-059 (Verbundvorhaben, Technologieförderung Land Mecklenburg-Vorpommern) in European Social / Regional Development Funds.

## B. Neural Network

### 1) Input

Our network processes entire writings without any further segmentation information. As it has been described in [2], the image is processed by reading its pixel data in four column-first "directions" that arise from combining top-down and bottom-up column traversals with left-to-right and right-to-left row traversals.

### 2) Structure

The layout of the recurrent neural network is copied from [2], but we use 50% more units in each hidden layer. In addition, the MDLSTM cells are substituted by MDLeaky cells. The motivation is that, for two-dimensional cell structures, the previously used MDLSTM cells may exhibit a certain instability by an exponential growth of their internal state. In contrast, the MDLeaky cells calculate their internal state as a convex combination of previous internal states and the new input. As a consequence, their maximal state is bounded by the maximal activation of their input. The MDLeaky cell has the same number of trainable weights as the MDLSTM cell. In various (non-public) handwriting recognition applications, we observed that MDLeaky cells outperform MDLSTM cells. This has been confirmed in our tests for the OpenHART 2013 DIR task where using MDLeaky cells instead of MDLSTM cells, the WER percentage decreases by approximately 1.0 for decoder #1 and approximately 0.5 for decoder #5 (see sections IV and V).

### 3) Output

The network has one output neuron for each character (class) under consideration and one extra "blank" channel for capturing special situations like transitions between characters. We simply worked over an alphabet of 152 characters appearing in the MADCAT data sets provided at the beginning of OpenHART 2013: 47 Arabic characters (mainly from U+0600 – U+06FF), 53 characters and the 10 digits from Latin character sets, and 42 special signs like e.g. brackets, quotes, commas, percent, space, … Due to the lack of specific knowledge of Arabic, we did not apply any preprocessing of the target string, i.e. we deal with special Arabic signs like vowels, hamzas and tatweels like ordinary characters. But we did introduce extra character classes for double and triple dots (ellipsis) because this turned out to be helpful with respect to the specific labeling we found in MADCAT references.

### 4) Training

The activation of an output neuron at a particular time is trained to estimate the probability of the occurrence of its corresponding character at a specific position. The network is trained by Backpropagation-Through-Time (BPTT) using the Connectionist Temporal Classification (CTC) algorithm described in [3] to calculate the gradient.

## C. Decoding

For constructing the hypothesis string for a given probability matrix of all output activations for the entire writing presented to the network, we compute the most likely character sequence arising from that matrix. With no restrictions that is just the sequence of the characters with the largest network output activation. If, otherwise, a dictionary is used to define permissible words, Viterbi decoding yields the one with highest probability.

As our dictionaries did not comprise sufficient language knowledge, we had to address the special case that all words of the dictionary are too unlikely. More specifically, if the average character probability over the best dictionary entry falls below a constant threshold θ, then we assume that the reference does not belong to the dictionary, and we take the most probable output sequence as the decoding result. By default, this threshold θ has been set to $1/e$.

Finally, in order to properly meet the MADCAT references, some decodings have to be partially reverted according to different Arabic and Latin writing's directionality. This has simply been decided by checking whether or not all letters of a certain word belong to Arabic unicode segments.

## D. Training environment

For train the ARGUS system, we took the MADCAT Phase 2 Training Set (LDC 2010 E17), thus using 27915 page images. One training epoch consists of presenting each of those images, but in order to keep training time realistic, only one, each time randomly chosen line of every page image has been used to train the network. That means, in each epoch, the network is trained on 27915 lines of different images. The images of MADCAT Phase 3 Training Set (LDC 2011 E97) were intentionally used to check for overfitting, but in fact, this had never been observed. Other OpenHART 2013 images of MADCAT Phase 1 Training or Evaluation (LDC 2010 E15, LDC 2008 E52) or OpenHART 2013 Dryrun (MADCAT Phase 2 Evaluation, LDC 2012 E52) have never been used for neither training nor validation. The BPTT network training is accomplished with a momentum factor of 0.9, and a learning rate decreasing gradually over the epochs as shown in Table I.

TABLE I. LEARNING RATES

| Learning rate | 1e-3 | 5e-4 | 2e-4 | 1e-4 | 5e-5 |
|---|---|---|---|---|---|
| Epochs | 1 .. 44 | 45 .. 60 | 61 .. 198 | 199 .. 228 | 229 .. 283 |

## IV. KEY DIFFERENCE IN CONTRASTIVE SYSTEMS

All CITlab systems use the very same preprocessing as described above, and the two network variations only differ in the final decrease of the learning rate: While the systems with network label *s* precisely follow Table I, the ones with network label *l* have been trained with learning rate $10^{-4}$ in epochs 199..276 and with learning rate $5 \cdot 10^{-5}$ in only one concluding epoch.

The systems' essential difference concerns the dictionaries: They simply consist of Arabic words which appear as tokens in selected MADCAT data sets provided for OpenHART 2013. For some systems we took all those words, for others only those which appear at least three times. The motivation is to prevent rare (faulty?) MADCAT tokens to become words in the dictionary, because we noticed that those might more mislead the dictionary lookup than that they help finding correct words. Finally, in one system, a larger value of $\theta = 1/\sqrt{e}$ was used (see III.B), that means, the raw reading result was preferred over the dictionary lookup optimum. The following Table II summarizes the respective decoder properties, where #1 is omitted because it does not use any dictionary at all.

TABLE II. DECODER SPECIFICATIONS

|  | Decoder | | | | |
| --- | --- | --- | --- | --- | --- |
|  | #2 | #3 | #4 | #5 | #6 |
| MADCAT Phase 1-3 Training Sets | yes | yes | yes | yes | yes |
| MADCAT Phase 1 Evaluation Set | yes | yes | yes | yes | yes |
| MADCAT Phase 2 Evaluation Set (OpenHART Dryrun) | no | no | no | yes | yes |
| only words with ≥ 3 occurences | no | no | yes | no | yes |
| enlarged θ | no | yes | no | no | no |

## V. RESULTS

The results were calculated with OpenHART's evaluation pipeline, version 1.1.2. The DIR results with CITlab's software ARGUS show that a well chosen decoder has more essential influence on the WER than the network itself. Although having trained and tested many networks with different parameters and/or initializations, we therefore finally decided to submit various decoding systems mainly using just one network. Also, we did not explore the capabilities of expert systems with more networks. Note that, while we declared ARGUS_s_5 as primary system, ARGUS_s_3 finally lead to our best result.

TABLE III. WORD ERROR RATES OF ALL CITLAB SUBMISSIONS

|  | ARGUS WER | |
| --- | --- | --- |
|  | *network s* | *network l* |
| **decoder #1** | 33.14 | 33.02 |
| **decoder #2** | 26.31 |  |
| **decoder #3** | <u>24.60</u> |  |
| **decoder #4** | 25.18 |  |
| **decoder #5** | **26.27** | 26.15 |
| **decoder #6** | 25.35 |  |

## VI. CONCLUSIONS

The OpenHART 2013 competition showed that CITlab's ARGUS software was able to deliver acceptable results on the overall DIR task. Originally intending to take part in a pure word recognition task, we learned only in the course of working on the more complex DIR task to deal with Arabic language and dictionary matters or OpenHART 2013 WER subtleties. As these were and are out of CITlab team's experience and core work area, only basic character sets and dictionaries have been used which were exclusively derived from the MADCAT data sets provided within OpenHART 2013. We believe that, basing on our network's raw reading output, further improvements of the overall WER are possible with more profound knowledge and exploitation of Arabic language matters as these can be much more essential for the overall performance than the neural network itself.

ACKNOWLEDGMENT

We are indebted to Alex Graves for several helpful discussions and comparisons with his previously used networks and earlier recognition results. Furthermore, the authors wish to thank the anonymous referees for their hints to and comments on an earlier version of this paper which lead to essential improvements.